\documentclass[10pt,twocolumn,letterpaper]{article}

\usepackage[pagenumbers]{cvpr} %

\usepackage{graphicx}
\usepackage{amsmath}
\usepackage{amssymb}
\usepackage{booktabs}
\usepackage{xcolor}
\usepackage{array,makecell}
\usepackage{caption}
\usepackage{graphicx}
\usepackage{wrapfig,lipsum}
\usepackage{enumitem}
\usepackage{amsmath}
\usepackage{bm}
\usepackage{multirow}
\usepackage[accsupp]{axessibility}

\usepackage[pagebackref,breaklinks,colorlinks]{hyperref}

\usepackage[capitalize]{cleveref}
\crefname{section}{Sec.}{Secs.}
\Crefname{section}{Section}{Sections}
\Crefname{table}{Table}{Tables}
\crefname{table}{Tab.}{Tabs.}

\newcommand{\vl}{{VL}}
\def\svlcfull{{Structured \vl{} Concept}}
\def\svlc{{SVLC}}

\newcommand{\oursbench}{ConStruct-VL} %
\newcommand{\oursbenchfull}{Continual Data-Free Structured VL Concepts Learning } %
\newcommand{\oursarch}{LaLo}
\newcommand{\oursarchfull}{Layered-LoRA}
\newcommand{\ourslossfull}{Adversarial Pseudo-Replay}
\newcommand{\oursloss}{APR}

\newcommand\secvspace{\vspace{-0.2cm}}
\newcommand\eqvspace{\vspace{-0.1cm}}

\newcommand\figvspace{\vspace{-0.5cm}}
\newcommand\tabvspace{\vspace{-0.6cm}}

\newcommand{\blu}[1]{{\color{blue}#1}}

\newcommand{\SupB}{Appendix Section \ref{appendix-prompt}}

\newcommand{\AppRefA}{Section~\ref{sec:exp} }
\newcommand{\AppRefB}{\cref{sec:exp}}
\newcommand{\AppRefC}{(\cref{tab:7task}, \cref{tab:7task-vg}, \cref{tab:4task-vaw})}

\makeatletter
\newcommand{\printfnsymbol}[1]{%
  \textsuperscript{\@fnsymbol{#1}}%
}
\makeatother

\begin{document}

\title{ConStruct-VL: Data-Free Continual Structured VL Concepts Learning}

\author{
\textbf{James Seale Smith\thanks{Equal contribution}\,\,\thanks{Work done during internship at MIT-IBM Watson AI Lab.}\,\,\textsuperscript{1,2}
\quad Paola Cascante-Bonilla\textsuperscript{1,3}
\quad Assaf Arbelle\textsuperscript{4}} \\ 
\textbf{Donghyun Kim\textsuperscript{1,4}
\quad Rameswar Panda\textsuperscript{1,4}
\quad David Cox\textsuperscript{1,4}
\quad Diyi Yang\textsuperscript{5}} \\
\textbf{Zsolt Kira\textsuperscript{2}
\quad Rogerio Feris\textsuperscript{1,4}
\quad Leonid Karlinsky\printfnsymbol{1}\textsuperscript{1,4}} \\
\\
\normalsize
\textsuperscript{1}MIT-IBM Watson AI Lab
\quad  \textsuperscript{2}Georgia Institute of Technology
\quad \textsuperscript{3}Rice University
\\
\normalsize
\quad \textsuperscript{4}IBM Research
\quad \textsuperscript{5}Stanford University}

\maketitle

\begin{abstract}
\secvspace
    \looseness=-1 Recently, large-scale pre-trained Vision-and-Language (\vl{}) foundation models have demonstrated remarkable capabilities in many zero-shot downstream tasks, achieving competitive results for recognizing objects defined by as little as short text prompts. However, it has also been shown that \vl{} models are still brittle in \svlcfull{} (\svlc{}) reasoning, such as the ability to recognize object attributes, states, and inter-object relations. This leads to reasoning mistakes, which need to be corrected as they occur by teaching \vl{} models the missing \svlc{} skills; often this must be done using private data where the issue was found, which naturally leads to a data-free continual (no task-id) \vl{} learning setting. In this work, we introduce the first \oursbenchfull{} (\oursbench{}) benchmark and show it is challenging for many existing data-free CL strategies. We, therefore, propose a data-free method comprised of a new approach of \ourslossfull{} (\oursloss{}) which generates adversarial reminders of past tasks from past task models. To use this method efficiently, we also propose a continual parameter-efficient \oursarchfull{} (\oursarch{}) neural architecture allowing no-memory-cost access to all past models at train time. We show this approach outperforms all data-free methods by as much as $\sim7$\% while even matching some levels of experience-replay (prohibitive for applications where data-privacy must be preserved). Our code is publicly available at~\url{https://github.com/jamessealesmith/ConStruct-VL}
    
\end{abstract}

\secvspace
\vspace{-0.4cm}
\section{Introduction}
\label{sec:intro}
\secvspace
\begin{figure}[t]
    \figvspace
    \centering
    \includegraphics[clip, trim=3.4cm 3.3cm 11cm 1.3cm, width=0.38\textwidth]{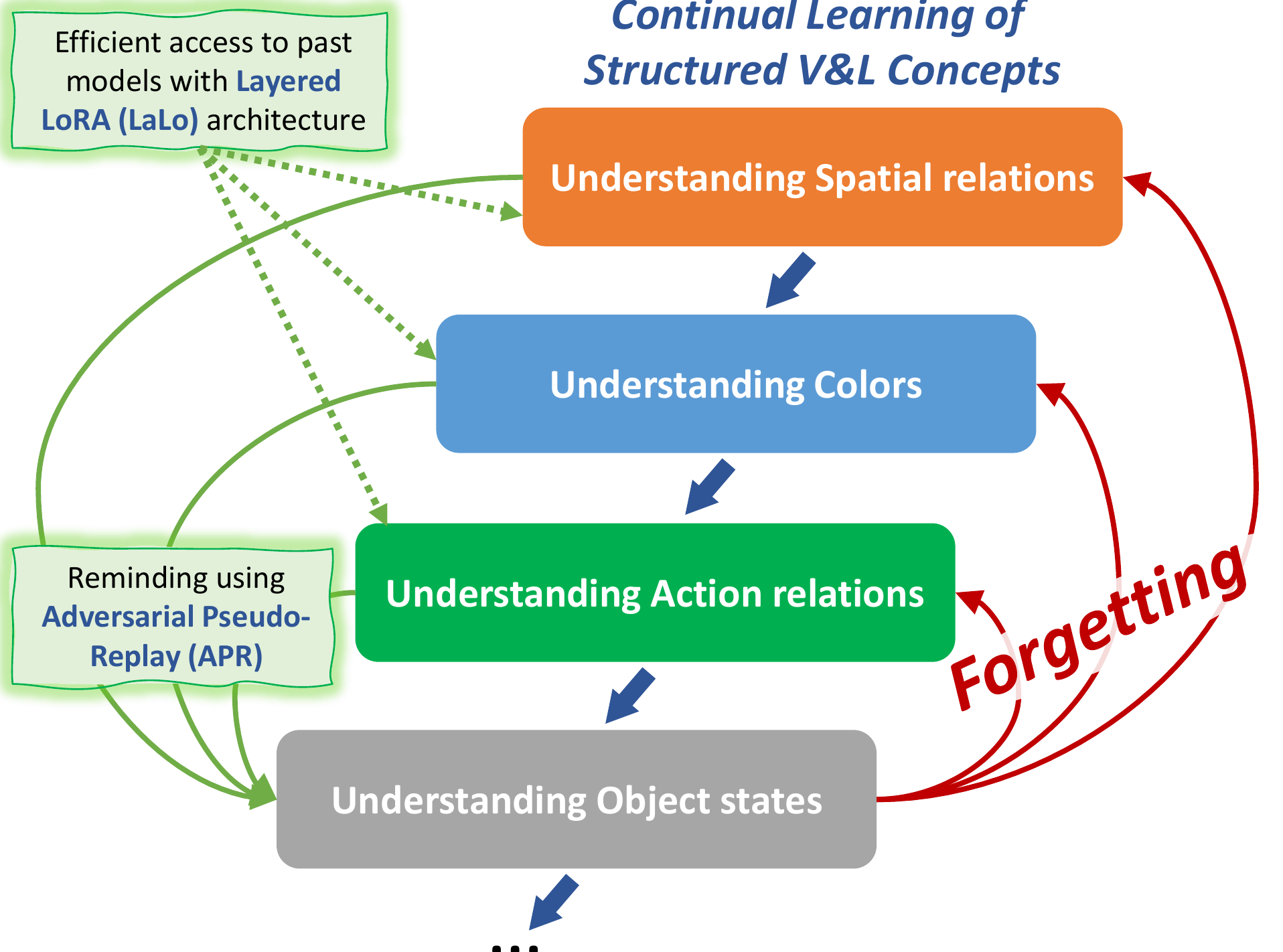}
    \vspace{-0.2cm}
    \caption{Illustration for the \oursbenchfull{} (\oursbench{}). \svlcfull{} (\svlc{}) understanding skills are added / refined over time, with \ourslossfull{} (\oursloss{}) effectively countering catastrophic forgetting using our \oursarchfull{} (\oursarch{}) architecture's ability of efficient no-memory-cost access to past task models.}
    \label{fig:concept}
    \vspace{-.5cm}
\end{figure}

\begin{figure*}[t]
    \figvspace
    \centering
    \includegraphics[clip, trim=0cm 4cm 0cm 3cm, width=0.85\textwidth]{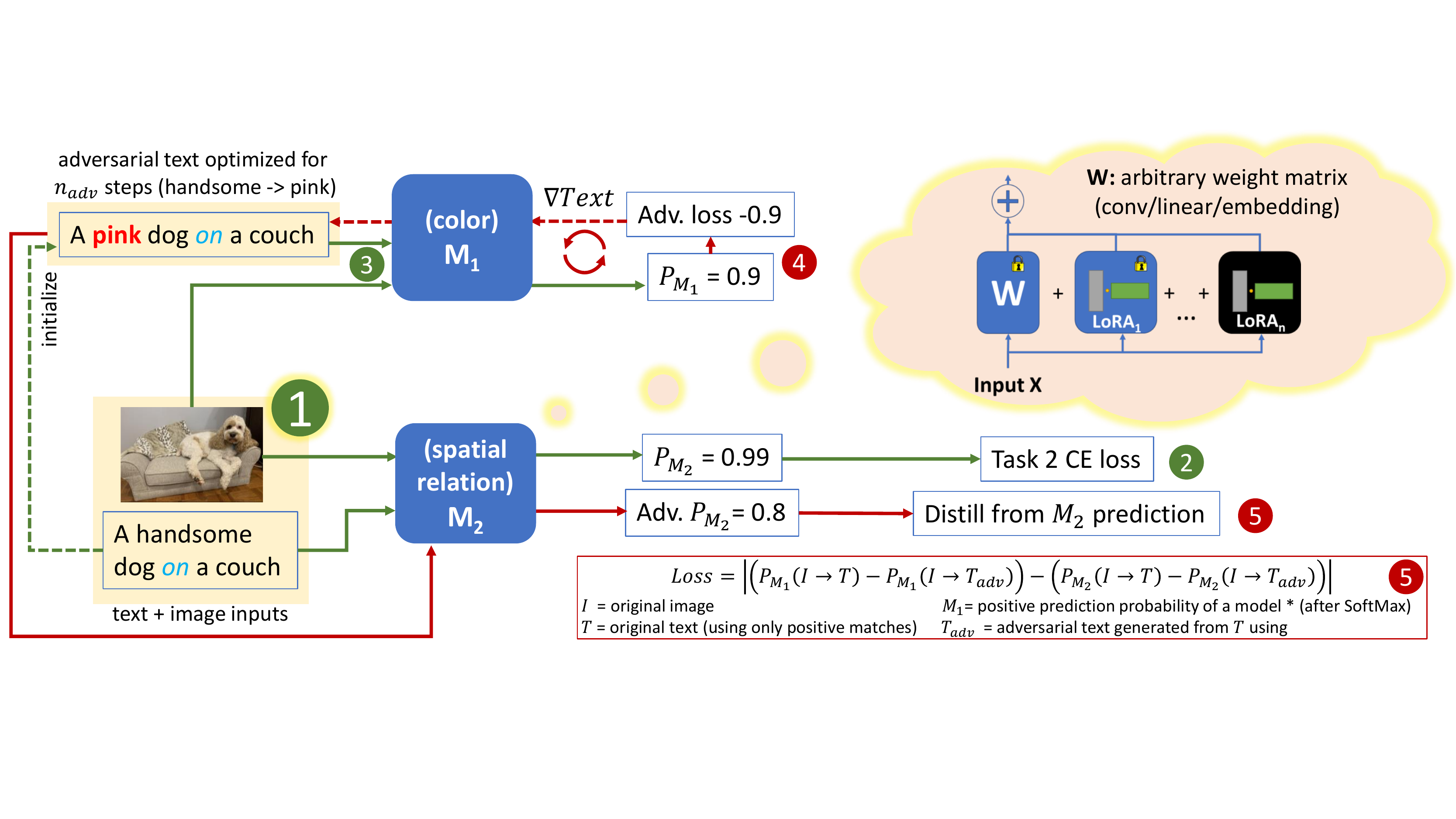}
    \vspace{-0.2cm}
    \caption{\oursarchfull{} (\oursarch{}) architecture and \ourslossfull{} (\oursloss{}), illustrated for a two-task \oursbench{} sequence teaching \textbf{`color'$\rightarrow$`spatial relation'} understanding skills. (1) A text+image pair is entering the model $M_2$ currently training to understand spatial relations; $M_2$ is an extension of the model $M_1$ (previously trained to understand color) via adding a layer of low-rank (LoRA) adapters to every parametric function of $M_1$ while keeping $M_1$ frozen. (2) $M_2$ produces positive prediction probability $P_{M_2}$ (after softmax) and applies CE loss. (3) Text+image pair are passed through $M_1$ (done without reloading in \oursarch{}). (4) $M_1$ produces a positive prediction probability $P_{M_1}$; using $-P_{M_1}$ as adversarial loss, we perform a sign-gradient attack on the text input (after tokenizer+first embedding layer). The produced text is expected to be adversarial in the direction of the \svlc{} used to train $M_1$, color in this case. We, therefore, expect some embedded text tokens to become `colored' inconsistently with the image. (5) We use the produced adversarial sample to distill from $M_1$ into $M_2$ via the proposed pseudo-replay positive prediction probability drop preserving loss.
    }
    \label{fig:intro}
    \figvspace
\end{figure*}

Recently, large Vision-and-Language (\vl{}) models achieved great advances in zero-shot learning \cite{clip,align,lxmert,uniter,oscar,vilt,align,tcl,blip}. Pre-trained on hundreds of millions \cite{clip} or billions \cite{laion} of image-and-text pairs collected from the web, these \vl{} models have demonstrated remarkable capabilities in understanding (recognizing \cite{clip,align}, detecting \cite{zareian2021open}, segmenting \cite{xu2021simple}, etc.) objects appearing in images or videos~\cite{qian2022multimodal}, thus moving beyond the previous paradigm of fixed object classes to open-vocabulary models.

Despite these great advances, several recent works \cite{winoground,vlc} have found \vl{} models to be brittle with respect to understanding \svlcfull{}s (\svlc{}s) - non-object textual content such as object attributes, states, inter-object relations (e.g. spatial, interaction, etc.), and more. 
Naturally, it is important to alleviate these shortcomings, as this lack of understanding of \svlc{}s can lead to embarrassing errors on the \vl{} model's part for many applications, such as analysis of multi-modal social networks data, multi-modal chat, multi-modal document analysis, and more. Importantly, in most of these applications: (i) different errors made by the model surface sequentially in the course of time, each time identifying another \svlc{} `skill' missing in the model; (ii) typically, to fix the errors, one would attempt additional fine-tuning of the model on a data collected from the source on which the errors were observed (hoping to avoid catastrophic forgetting of the previous \textit{model improvement} rounds); (iii) data sources where errors are observed are typically private and cannot be passed between these fine-tuning tasks. 
We, therefore, ask the question: Can we sequentially add \svlc{} skills to multi-modal \vl{} models, in a privacy-preserving manner? This leads to the
\textit{data-free continual learning problem} (\eg, \cite{l2p,dualprompt}) cast in the multi-modal \vl{} domain.

To this end, we introduce the first \emph{\oursbenchfull{} (\oursbench{}) multi-modal benchmark}, built on top of the popular Visual Genome \cite{vg} and Visual Attributes in the Wild \cite{vaw} datasets using the protocol proposed in VL-Checklist \cite{vlc}, and show it is challenging for many existing data-free CL strategies, including recent SOTA prompting-based CL methods~\cite{l2p,dualprompt}. We then offer a novel data-free CL method, leveraging the multi-modal nature of the problem, to effectively avoid forgetting.%
We propose the concept of \ourslossfull{} (\oursloss{}), that (as opposed to the previous pseudo-replay works~\cite{choi2021dual,smith2021abd,yin2020dreaming}) generates \textit{negative} examples to past task models conditioned on the current batch data. For continual \vl{} training we generate negatives in one of the modalities by making it inconsistent with the other modality via a series of adversarial attack steps utilizing past models (see Fig.~\ref{fig:intro} for an example). Intuitively, generating (inconsistent) negatives (\oursloss{}) is easier than generating (consistent) positives (pseudo-replay). Also, past task model attacks are likely to generate inconsistencies corresponding to their tasks, thus leading to reduced forgetting when we use the generated negative samples 
to distill from the past task models the drop in prediction probabilities after adversarial examples are applied
(\cref{fig:concept}). To use the proposed \oursloss{} technique we need the ability to efficiently invoke past models at training time. We, therefore, propose a \oursarchfull{} (\oursarch{}) continual learning neural architecture utilizing layered parameter-efficient (low-rank) residual adapters, supporting invocation of any of the past task models on any given training batch at no additional memory cost (without the need to reload these past models into memory). Moreover, our proposed architecture can be collapsed to the original model size by collapsing all the adapters into their adapted parametric functions, thus supporting inference on the final model at no additional cost. 

\noindent\textbf{Contributions:} (i) we propose the challenging \oursbench{} benchmark,
and show that existing data-free CL methods struggle in this setting; (ii) we propose a new concept of \ourslossfull{} (\oursloss{}) specifically designed for multi-modal continual learning, alongside a \oursarchfull{} (\oursarch{}) architecture allowing invoking any of the past task models efficiently without reloading and having no additional inference time or parameters cost; and (iii) we demonstrate significant improvements (over $6.8\%$ increase in final accuracy and $\times5$ smaller average forgetting) of the proposed approach compared to all the popular data-free CL baselines, as well as some amounts of experience replay.

\secvspace
\vspace{-0.1cm}
\section{Related Work}
\label{sec:rl}
\secvspace

\noindent\textbf{Vision-and-Language (\vl{}) Models.} \vl{} models pre-trained on large-scale noisy image-text data with contrastive loss (\eg, CLIP~\cite{clip} and ALIGN~\cite{align}) show remarkable performance in many zero-shot downstream tasks. Some of the methods improve image-text alignment by focusing on region features from the off-shelf object detectors~\cite{lxmert,uniter,oscar} or use end-to-end cross-attention layers with additional objective functions such as image-text matching and masked language modeling~\cite{vilt,albef,tcl,blip}, or filtering noisy captions (\eg, BLIP~\cite{blip}). Some \vl{} models~\cite{cyclip,filip,cloob,declip,pyramidclip} utilize additional properties of language structure. DeCLIP~\cite{declip} uses textual nearest-neighbor to find additional positives. CyClip~\cite{cyclip} learns geometrically consistent representations in the image and text embedding space. However, recent studies~\cite{vlc,winoground} have demonstrated the limitation of all of these models in learning \svlc{} reasoning skills. Even the most recent and sophisticated \vl{} methods like CyClip~\cite{cyclip} still struggles with \svlc{} understanding (\cref{tab:7task}). In our work, we study ways for continual improvement of \vl{} models in \svlc{} reasoning tasks and propose the first \oursbenchfull{} benchmark to facilitate this study in future works. 

\noindent\textbf{Continual Learning.} Continual learning aims to continuously adapt a model to non-stationary data distributions. 
Prior works can be categorized by their method for avoiding catastrophic forgetting~\cite{mccloskey1989catastrophic}. Regularization-based methods (\eg,~\cite{li2016learning, zenke2017continual, kirkpatrick2017overcoming,jung2016less}) introduce extra regularization terms in the objective functions when learning a new task. LwF~\cite{li2016learning} uses soft labels for the previous task to avoid forgetting. EWC~\cite{kirkpatrick2017overcoming} estimates the importance of parameters and uses this for per-parameter weight decay. Rehearsal-based methods (\eg,~\cite{Rebuffi:2016,chaudhry2018efficient,chaudhry2019episodic,hou2019learning,rolnick2019experience, kamra2017deep,ostapenko2019learning,van2020brain,pham2021dualnet}) use a data buffer to store or generate some samples of the previous tasks. These samples are replayed with new task data. However, in many cases storing samples may not be allowed due to privacy or copyright concerns. Architecture-based methods (\eg,~\cite{Rusu:2016,yoon2017lifelong,aljundi2017expert,li2019learn,l2p,dualprompt}) isolate model parameters for each task. Recently, prompt-based continual learning methods (\eg, L2P~\cite{l2p}, DualPrompt~\cite{dualprompt}, 
S-Prompt\footnote{S-Prompt is designed for a different setting than ours.}~\cite{wang2022sprompt}) outperformed  rehearsal-based methods without using a replay buffer.
Additional detailed related work can be found in~\cite{de2021continual}. All of the above methods target uni-modal (e.g. vision only) CL setting, while we focus on multi-modal \vl{} setting in this work.  REMIND~\cite{hayes2020remind} proposed Continual VQA tasks with latent replay, but unlike ours requires storing compressed training data. CLiMB \cite{climb} proposed CL adaptation to coarsely different \vl{} tasks, such as VQA, NLVR, SNLI-VE, VCR, and assumed knowledge of the evaluated task-id at inference time. In contrast, we target continual adaptation to a fine-grained sequence of \svlc{} reasoning skills, with no task-id knowledge during inference.%

\noindent\textbf{Parameter-Efficient Adaptation.} 
Several approaches~\cite{ding2022delta,mao2021unipelt,vladapter} have been proposed for efficient model fine-tuning using fewer parameters. Either fine-tuning a subset of the model~\cite{zaken2021bitfit,sung2021training}, adding adapters~\cite{houlsby2019parameter,pfeiffer2020adapterfusion,rebuffi2017learning}, low-rank adapters (LoRA~\cite{lora}), or learning prompts~\cite{li2021prefix,zhou2022learning,lester2021power}. UNIPELT~\cite{mao2021unipelt} proposes a unified framework for NLP that subsumes the method of adapters, prompt tuning, and LoRA with gating mechanism. VL-ADAPTER~\cite{vladapter} presents a unified multi-task learning framework for \vl{} tasks evaluating the use of adapters, Hyperformer~\cite{mahabadi2021parameter}, and Compactor~\cite{zhang2021beyond} for \vl{} tasks. In this work, we propose an efficient CL neural architecture built on the concept of layering low-rank adapters on a \vl{} model effectively allowing replaying past task models at no additional memory cost.

\secvspace
\section{Method}
\label{sec:method}

\secvspace
\subsection{Tasks} 
\secvspace
In this work we target adapting \vl{} models to an arbitrary sequence of \svlc{} reasoning tasks. Recently, several datasets (e.g., \cite{winoground} and \cite{vlc}) have been proposed for these tasks, given their importance. Each task $\mathcal{T}^i$ is defined by a set of image $I$ and text $T$ pairs, and a binary ground-truth function $\psi^i(T,I) \in \{0,1\}$ indicating whether the text of the pair corresponds to the image content or not (\cref{fig:vlc_examples}a). The difference between the tasks is the type of \svlc{} that they are `teaching'. For example, there could be tasks teaching the understanding of color, material, size, state, or other object attributes by explicitly having words related to these attributes contained inside the task's texts $T$ (\cref{fig:vlc_examples}a). Other tasks could teach skills for understanding positional or transitive action relations between multiple objects being present in the image and interacting with each other. In practical applications, tasks could teach understanding the presence of unwanted attributes or relations, such as nudity, profanity, violence, etc. Alternatively, they could teach understanding positive, but personal, aspects including love, affection, or any type of personal preference from private captioned photo collections. As their nature, these types of task data (both images and text) could be very sensitive and private information, prohibiting maintaining any of these task's data between tasks training (e.g. tasks data could be hosted by different institutions, or privately belong to different users of a social network). This naturally leads to the \textit{continual data-free setting with no task-id}, 
which is exactly the setting on which we focus. Note that methods that require detecting to which task a given input belongs at inference time are not reasonable here, as it is difficult to estimate what \svlc{} skill is needed from free-form text and a mixture of skills may be needed.%

\secvspace
\subsection{Model} 
\secvspace

Our model architecture is based on BLIP \cite{blip} (an openly available VLP with strongest support for \svlc{} understanding as we show in \cref{tab:7task}). BLIP is comprised of separate text and image encoders followed by a cross-attention decoder admitting both streams and followed by a classifier (for \svlc{} reasoning). Thanks to its modular architecture, the model supports both zero-shot retrieval and \svlc{} reasoning tasks. Our proposed method of model adaptation by sequentially improving the model's \svlc{} understanding skills maintains this support leaving the zero-shot retrieval pipeline intact and unchanged. 

\secvspace
\subsubsection{\oursarchfull{} (\oursarch{}) Architecture}\label{sec:arch}
\secvspace
Our model admits a text-and-image pair $(T,I)$ and is comprised of four parts: (i) image encoder $e_I = \mathcal{E}_I(I)$; (ii) text encoder $e_T = \mathcal{E}_T(T)$; (iii) cross-attention decoder $d_{I,T} = \mathcal{D}(e_T, e_I)$; and (iv) binary classification head $\mathcal{C}(d_{I,T}) \in \{0,1\}$. The final model is, therefore:
\begin{equation}
    \eqvspace
    \mathcal{M}(T,I) = \mathcal{C}(\mathcal{D}(\mathcal{E}_T(T), \mathcal{E}_I(I)))
    \eqvspace
\end{equation}
The decoder $\mathcal{D}$ is an extension of the text encoder $\mathcal{E}_T$, with cross-attention blocks inserted between each two consecutive self-attention blocks as well as self-attention block parameters being shared between $\mathcal{D}$ and $\mathcal{E}_T$. Each of the cross-attention blocks of $\mathcal{D}$ also admits the embedded image tokens $e_I$ as additional hidden states and employs them in the standard cross-attention computation of the prefix-tuning type (meaning output size is the same as input, and only hidden states of the text tokens are passed between the blocks). Each of the $\mathcal{E}_T$, $\mathcal{E}_I$, and $\mathcal{D}$ networks is comprised of a mix of non-parametric functions (here referring to any kinds of data norms, e.g. \textit{LayerNorm}, as non-parametric since we have them frozen) and two types of parametric functions: linear (e.g. \textit{torch.nn.Linear} in PyTorch \cite{pytorch}) and embedding (e.g. \textit{torch.nn.Embedding}). Roughly, each of those functions $\mathcal{F}^*_k(x)$ (where $*$ is either $lin$ or $emb$) is parameterized by a weight matrix $\mathcal{W}_k$ so that:
\begin{align}
\eqvspace
    &\mathcal{F}^{lin}_k(x)=\mathcal{W}_k \cdot x \\
    &\mathcal{F}^{emb}_k(x)=EMB(x;\mathcal{W}_k)
\eqvspace
\end{align}
where $EMB$ is the embedding operator picking the respective columns of $\mathcal{W}_k$ assuming $x$ is a stream of integers. Also, we disregard possible bias terms of the linear functions, as if they are present we keep them frozen. Assuming a sequence of $n$ tasks $[\mathcal{T}^i]_{i=1}^n$, we train a sequence of models $[\mathcal{M}^i]_{i=1}^n$ where the weights $\mathcal{W}_k^i$ of the model $\mathcal{M}^i$ employed in the task $\mathcal{T}^i$ training are parameterized as:
\begin{equation}
\eqvspace
    \mathcal{W}_k^i = \mathcal{W}_k^{i-1} + \mathcal{A}_k^i \cdot \mathcal{B}_k^i
\eqvspace
\end{equation}
where $\mathcal{W}_k^{0}$ are the weights of the base model before training any task, and, following the idea proposed in \cite{lora} for efficient LLM fine-tuning using Low-rank Residual Adapters (LoRA), $\mathcal{A}_k^i$ and $\mathcal{B}_k^i$ are (low) rank $r$ matrices of sizes $m \times r$ and $r \times l$ respectively (with $m \times l$ being the dimensions of $\mathcal{W}_k^i$). These LoRA adapters can be applied efficiently as:
\begin{align}
\eqvspace
    &\mathcal{F}^{i,lin}_k(x)=\mathcal{F}^{i-1,lin}_k(x) + \mathcal{A}_k^i \cdot (\mathcal{B}_k^i \cdot x) \\
    &\mathcal{F}^{i,emb}_k(x)=\mathcal{F}^{i-1,emb}_k(x) + \mathcal{A}_k^i \cdot EMB(x;\mathcal{B}_k^i)
\eqvspace
\end{align}
During the training of task $\mathcal{T}^i$, we freeze all previous tasks' (and base's) model parameters: $\forall k,\{\mathcal{W}_k^j\}_{j=0}^{i-1}$, and only learn current task LoRA adapters: $\forall k,\{(\mathcal{A}_k^i,\mathcal{B}_k^i)\}$. 

\looseness=-1
There are several interesting things to note about the proposed \oursarch{} architecture: 
(i) as opposed to \cite{vladapter}, who evaluated the use of LoRA for (single task) \vl{} models finding it ineffective, we also add our LoRA adapters to all the layers of the image encoder, and not only to the text encoder/decoder as done in \cite{vladapter}; 
(ii) as opposed to \cite{sidetuning} who attach a small side network only to the \textit{output} of the adapted model (also requiring task-id in continual setting), our LoRA adapters are added to all parametric functions inside the model and affect all the intermediate computations; (iii) at the end of training, we expect our last model $\mathcal{M}^n$ to have acquired all the \svlc{} understanding skills taught by the task sequence $[\mathcal{T}^i]_{i=1}^n$ with minimal forgetting (\cref{sec:forgetting}), and hence, at inference, all the LoRA adapters can be folded back into the original weight matrices by simple summation leading to \textit{zero} additional inference cost; (iv) with rank $r$ kept low, the number of extra parameters added by all the LoRA adapters is very low (e.g. $2.8\%$ in our experiments); finally, (v) we keep our classifier $\mathcal{C}^i=\mathcal{C}$ shared across all task models - in practice, we only train it for the first task $\mathcal{T}^1$ and keep it frozen for all later tasks - this reduces some model capacity and could be improved in future work. 

\secvspace
\subsubsection{\ourslossfull{} (\oursloss{})}\label{sec:forgetting}
\secvspace

One of the benefits of our proposed \oursarch{} architecture is the ability to efficiently invoke any past tasks model $\mathcal{M}^i$ by stopping the iterative (on $j$) computation of $\mathcal{F}^{j,*}_k(x)$ at $j=i$. This gives us an effective tool for battling forgetting by allowing us to revisit any past model as part of a single training batch computation without the need to reload any weights (infeasible for modern model sizes). We use this neat potential for our data-free \oursloss{} multi-modal continual learning strategy to make our current model $\mathcal{M}^i$ training on task $\mathcal{T}^i$ not forget the `lessons' previously `taught' by tasks $\{\mathcal{T}^j\}_{j=1}^{i-1}$. In \oursloss{}, we use task $\mathcal{T}^i$ positive $(T,I)$ pairs ($\psi^i(T,I)=1$) for adversarially simulating negatives with respect to past task models, and enforcing a loss requiring the current model's positive prediction probability to drop in the presence of the simulated negatives by a similar amount as it would drop for the past task model.%

Next, we describe how to simulate a negative sample for a past task model $\mathcal{M}^j=(\mathcal{E}_T^j,\mathcal{E}_I^j,\mathcal{D}^j,\mathcal{C})$, with $j<i$, given a text and image pair $(T,I) \in \mathcal{T}^i$, s.t. $\psi^i(T,I)=1$. Let the adversarial gradient be defined as:
\begin{equation}
\eqvspace
    \nabla^j_*(T,I) = - \frac{\partial}{\partial *} POS(\mathcal{C}(\mathcal{D}^j(\mathcal{E}_T^j(T), \mathcal{E}_I^j(I))))
\eqvspace
\end{equation}
where $POS$ is the positive prediction probability of the classifier $\mathcal{C}$ \textit{following softmax} (taking it before softmax would not regularize the negative prediction probability), and $*$ can be either image $I$ or text $T$ (since $T$ is a discrete output of a tokenizer, when computing adversarial samples using text we freeze the text embedding layer for all models and compute adversarial gradients with respect to its continuous output).  Without loss of generality, we describe how to obtain an adversarial text $T_{adv}^j$ with respect to a past model $\mathcal{M}^j$ from a text $T$ corresponding image $I$ in task $\mathcal{T}^i$. Similarly, we could compute an adversarial image $I_{adv}^j$. In practice, we use adversarial text in all of our experiments, which leads to less forgetting and better performance (\cref{sec:ablation}). We initialize $T_{adv}^j=T$ and update $T_{adv}^j$ in each adversarial step via the sign-gradient adversarial attack: 
\begin{equation}
\eqvspace
    T_{adv}^j \leftarrow T_{adv}^j + \lambda_{adv} \cdot sign(\nabla^j_T(T_{adv}^j,I))
\eqvspace
\end{equation}
where $\lambda_{adv}$ is the step size, and we use $n_{adv}$ update steps. Note that this process is multi-modal as the adversarial text $T_{adv}^j$ is computed with respect to the image $I$, originally a positive pair of the starting text $T$. Some qualitative examples of generated $T_{adv}^j$ are in \cref{fig:vlc_examples}b. We compute all the $T_{adv}^j$ for all positive pairs in the training batch simultaneously. Having obtained the adversarial texts, we define the following \oursloss{} loss $\mathcal{L}_{adv}^j$ w.r.t. any past model $j<i$ as:
\begin{footnotesize}
\begin{align}
\eqvspace
    &\delta^k(T, T_{adv}^j, I) = POS(\mathcal{M}^k(T,I)) - POS(\mathcal{M}^k(T_{adv}^j,I))\\
    &\mathcal{L}_{adv}^j = AVG_{(T,I)}|\delta^j(T, T_{adv}^j, I) - \delta^i(T, T_{adv}^j, I)|
    \label{eq:adv_loss}
\eqvspace
\end{align}
\end{footnotesize}
where $POS$ is the positive prediction probability of the classifier head $\mathcal{C}$ (after softmax), and $AVG_*$ computes average over $*$. 
The intuition behind $\mathcal{L}_{adv}^j$ is that $T_{adv}^j$ will move the text $T$ to contain words corresponding to \svlc{} taught by the past task $\mathcal{T}^j$, but not present on the image $I$. For example, a text describing a `handsome dog on a couch' as part of \textit{spatial relations} training task, might be modified to include the color `pink' as adversarial modification by the past task model that has been taught the \textit{color} \svlc{} (\cref{fig:intro}). Maintaining similar response to this modification in the text between the normalized prediction probabilities of the current model $\mathcal{M}^i$ and the normalized prediction probabilities of the past model $\mathcal{M}^j$ will likely reduce forgetting, which is indeed verified by the strong effects of this loss in our experiments and ablations. An additional intuition here is that \oursloss{} is easier to do than positive pseudo-replay proposed in past works for the uni-modal continual setting~\cite{choi2021dual,smith2021abd,yin2020dreaming}. Indeed, producing adversarial effects, that is modifying the text to make it reference things not on the image, is easier than modifying the text to contain words corresponding to past task, but actually appearing on the image. This is verified in our ablations (\cref{sec:ablation}).

Finally, combined with the standard cross-entropy loss for the task $\mathcal{T}^i$ with respect to its ground truth annotations $\psi^i(T,I)$ we arrive at the final loss $\mathcal{L}^i$ used for training $\mathcal{M}^i$:
\begin{footnotesize}
\begin{equation}
\eqvspace
    \mathcal{L}^i = AVG_{(T,I)}(CE(\mathcal{M}^i(T,I),\psi^i(T,I))) + \rho \cdot \sum_{j=0}^{i-1} \mathcal{L}_{adv}^j
    \label{eq:final_loss}
\eqvspace
\end{equation}
\end{footnotesize}
where $\rho$ is a parameter. In our experiments we also evaluate a simplified variant of this loss where we use only the adversarial loss w.r.p.t. the previous model $\mathcal{L}_{adv}^{i-1}$ instead of the last summation term in \cref{eq:final_loss}.

\secvspace
\subsection{\oursbench{} Benchmark}\label{sec:bench}
\secvspace
\begin{figure}[t]
    \figvspace
    \centering
    \includegraphics[width=0.4\textwidth]{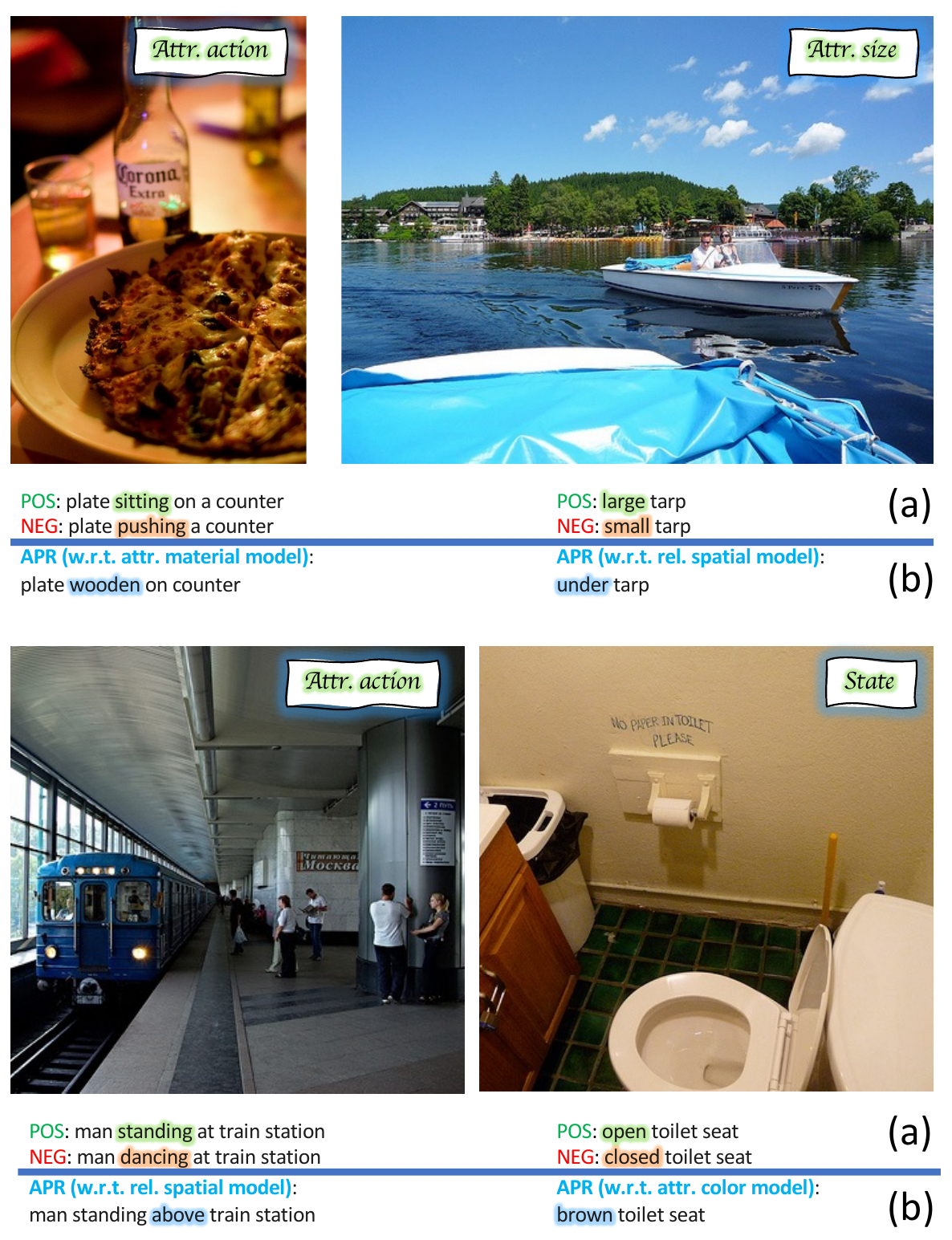}
    \vspace{-0.3cm}
    \caption{Some examples of the \oursbench{} tasks data and negatives generated by our proposed \oursloss{} technique w.r.t. past task models. (a) Showing image + POS text + NEG text triplets for \textit{(intransitive) action}, \textit{size}, and \textit{state} tasks; (b) Showing the adversarial text $T_{adv}$ generated w.r.t. previously trained \textit{material}, \textit{spatial relations}, or \textit{color} models, modifications highlighted}
    \label{fig:vlc_examples}
    \figvspace
    \vspace{-0.1cm}
\end{figure}
To the best of our knowledge, no prior works have proposed a benchmark to evaluate the continual adaptation of \vl{} models, teaching them reasoning over \svlc{}s (such as understanding of attributes, relations, and states) in a continual manner without task-id knowledge at test time (as we want the final model to understand all the concepts taught without informing it on which concept it is evaluated). Therefore, we propose the first such \oursbenchfull (\oursbench) benchmark. We construct \oursbench{} from the two popular publicly available \vl{} datasets: Visual Genome (VG) \cite{vg} and Visual Attributes in the Wild (VAW) \cite{vaw}, building on protocols proposed in the VL-Checklist \cite{vlc}. 

Specifically, we use $7$ \svlc{} subsets defined by \cite{vlc} for measuring \vl{} models' performance on understanding relations (spatial, inter-object transitive actions), attributes (size, color, material, intransitive single object actions), and object state. Each subset is constructed from triplets including a single image and two texts, one of them corresponding to the image content (positive) and another non-correspondent (negative) - manipulated by changing the value of a single \svlc{} (a specific concept for each subset). Some examples of the triplets for different subsets are provided in \cref{fig:vlc_examples}a. The subsets are separately constructed from each of the VG and VAW datasets image-text pairs, with all $7$ \svlc{} subsets available for VG and only the $5$ attribute and object state subsets defined for VAW. The sizes of the subsets range between $1$K to $31$K in VG and $3$K to $64$K in VAW. In addition to VG and VAW, we also define the combined VG+VAW dataset uniting the corresponding subsets of the same nature. The resulting VG+VAW dataset has the aforementioned $7$ subsets with sizes ranging between $5$K to $75$K. We split each subset to train / val / test according to the $80$ / $10$ / $10$ percent rule guaranteeing no overlap between the splits. In \oursbench{} for each of the 3 datasets (VG, VAW, VG+VAW) each subset is treated as a separate continual learning task, thus forming $7$ total available tasks used to construct the task sequences in our experiments.

\secvspace
\subsection{Baselines}\label{sec:baselines}
\secvspace

For a fair comparison, we compare to the state-of-the-art \emph{data-free} continual learning baselines \emph{which do not rely on task id}. These methods are: Learning-without-Forgetting (LwF)~\cite{li2016learning}, a method which distills soft model predictions from a prior, frozen model copy to the current model; Elastic-Weight-Consolidation (EWC)~\cite{kirkpatrick2017overcoming}, a method which penalizes changes to model parameters weighted by an importance matrix\footnote{The importance matrix is of the same dimension of the model parameter weights and is calculated using the diagonal terms of the Fisher-Information matrix w.r.t. past task loss functions and data.}; and L2, which is EWC without the importance-weighting matrix (which can sometimes outperform EWC in the presence of pre-training~\cite{smith2022closer}).

We also compare to state-of-the-art \emph{prompting for continual learning} methods L2P~\cite{l2p} and DualPrompt~\cite{dualprompt}. These approaches continuously learn a set of prompts which are matched with input data in an instance-wise fashion. To our surprise, we found that these methods struggled in our benchmark, despite a heavy search for prompt depth and length. Our intuition is that these methods rely on encoder knowledge that is strongly related to the continual learning task sequence and, unlike image classification, our proposed task sequences requires substantial semantic modifications to the transformer encoders.

Finally, we also compare to continual fine-tuning of the full pre-trained model over the task sequence using a small learning rate (referred to as CFT). CFT was found to be more effective than the prompting methods L2P/DualPrompt. We also compare to two variants of CFT: (i) CFT Frozen Head (CFT-FH), where the classifier head is frozen after task 1 (as done in our approach); and (ii) CFT Linear Probing (CFT-LP), where the encoders are frozen and only the classifier head is learned. %

\secvspace
\section{Experiments} \label{sec:exp}
\begin{table}[t]
\caption{\textbf{Results (\%) on 7 Task VG+VAW}. Fin. Acc. is the no-task-id performance of the final model on all tasks jointly; $A_N$ is the accuracy averaged over tasks; $F_N$ is the average forgetting; and $N_{prm}$ is the \% of trainable parameters. For L2P \cite{l2p} and DP \cite{dualprompt} we swept over multiple \% of $N_{prm}$ and report the highest result.\\ $^\dagger$ \textit{measured using original publicly available codes + weights}}
\label{tab:7task}
\centering
\vspace{-2mm}
\resizebox{.45\textwidth}{!}{
\begin{tabular}{c|l|c|c|c} 
\toprule
& \rule{0pt}{10pt} Method ($N_{prm}$) & Fin. Acc. ($\uparrow$) & $A_N$ ($\uparrow$) & $F_N$ ($\downarrow$)\\ %
\midrule
\multirow{8}{*}{\rotatebox[origin=c]{90}{VLPs}} &
BLIP$^\dagger$ \cite{blip} & $ 77.8 $ & $ 72.33 $ & $ - $ \\ %
& CLIP$^\dagger$ \cite{clip} & $ 62.99 $ & $ 59.93 $ & $ - $ \\ %
& CyCLIP$^\dagger$ \cite{cyclip} & $ 57.76 $ & $ 56.36 $ & $ - $ \\ %
& ALBEF$^\dagger$ \cite{albef} & $ 71.2 $ & $ 67.9 $ & $ - $ \\ %
& OSCAR$^\dagger$ \cite{oscar} & $ 71.04 $ & $ 67.69 $ & $ - $ \\ %
& ViLT$^\dagger$ \cite{vilt} & $ 70.35 $ & $ 68.41 $ & $ - $ \\ %
& UNITER$^\dagger$ \cite{uniter} & $ 67.23 $ & $ 65.29 $ & $ - $ \\ %
& LXMERT$^\dagger$ \cite{lxmert} & $ 68 $ & $ 64.39 $ & $ - $ \\ %
& TCL$^\dagger$ \cite{tcl} & $ 70.29 $ & $ 65.42 $ & $ - $ \\ %
\hline
\multirow{8}{*}{\rotatebox[origin=c]{90}{Baselines}} &
CFT (\small{100\%}) & $ 73.91 $ & $ 80.76 $ & $ 6.94 $ \\ %
& CFT-FH (\small{99.7\%}) & $ 77.43 $ & $ 83.73 $ & $ 6.14 $ \\ %
& CFT-LP (\small{0.3\%}) & $ 64.85 $ & $ 71.32 $ & $ 12.26 $ \\ %
& L2 (\small{100\%})  & $ 76.36 $ & $ 85.59 $ & $ 4.57 $ \\ %
& LwF (\small{100\%})  & $ 79.39 $ & $ 86.11 $ & $ 4.28 $ \\ %
& EWC (\small{100\%})  & $ 73.29 $ & $ 81.91 $ & $ 6.13 $ \\ %
& L2P \cite{l2p} (\small{$\le5\%$})  & $ 66.96 $ & $ 76.46 $ & $ 6.7 $\\ %
& DP \cite{dualprompt} (\small{$\le5\%$})  & $ 58.59 $ & $ 68.65 $ & $ 9.52 $ \\ %
\hline
\multirow{3}{*}{\rotatebox[origin=c]{90}{\textbf{Ours}}} &
Ours$^{rand}$ (\small{2.8\%})  & $ 78.6 $ & $ 86.19 $ & $ 1.5 $ \\ %
& Ours$^{i-1}$  (\small{2.8\%})  & $ \textbf{86.2} $ & $ \blu{90.37} $ & $ \blu{1.4} $ \\ %
& Ours (\small{2.8\%})  & $ \blu{85.4} $ & $ \textbf{90.88} $ & $ \textbf{0.75} $ \\ %

\bottomrule
\end{tabular}
}
\tabvspace
\end{table}
\begin{table*}[t]
\vspace{-0.3cm}
\caption{\textbf{Results (\%) on 7 Task VG and 4 Task VAW}. Fin. Acc. is the no-task-id performance of the final model on all tasks jointly; $A_N$ is the accuracy averaged over tasks; $F_N$ is the average forgetting; and $N_{prm}$ is the \% of trainable parameters. For L2P \cite{l2p} and DP \cite{dualprompt} we swept over multiple \% of $N_{prm}$ and report the highest result.}
\vspace{-3mm}
\centering
\begin{subtable}[h]{0.45\textwidth}
\centering
\caption{7 Task VG}
\label{tab:7task-vg}
\resizebox{0.9\textwidth}{!}{
\begin{tabular}{c|l|c|c|c} %
\toprule
& \rule{0pt}{10pt} Method ($N_{prm}$) & Fin. Acc. ($\uparrow$) & $A_N$ ($\uparrow$) & $F_N$ ($\downarrow$) \\
\midrule
\multirow{1}{*}{\rotatebox[origin=c]{90}{VLPs}} & 
\multirow{2}{*}{BLIP \cite{blip}} & \multirow{2}{*}{$ 79.58 $} & \multirow{2}{*}{$ 74.24 $} & \multirow{2}{*}{$ - $} \\
& & &  \\
\hline
\multirow{8}{*}{\rotatebox[origin=c]{90}{Baselines}} &
CFT (\small{100\%})  & $ 77.99 $ & $ 83.5 $ & $ 6.94 $ \\
& CFT-FH (\small{99.7\%})  & $ 81.30 $ & $ 85.51 $ & $ 5.88 $ \\
& CFT-LP (\small{0.3\%})  & $ 65.13 $ & $ 72.70 $ & $ 11.73 $ \\
& L2 (\small{100\%})   & $ 77.40 $ & $ 87.12 $ & $ 5.21 $ \\
& LwF (\small{100\%})   & $ 77.01 $ & $ 85.24 $ & $ 5.80 $ \\
& EWC (\small{100\%})   & $ 72.78 $ & $ 82.61 $ & $ 7.23 $ \\
& L2P (\small{$\le5\%$})   & $ 60.88 $ & $ 69.23 $ & $ 17.07 $ \\
& DP (\small{$\le5\%$})   & $ 65.55 $ & $ 66.94 $ & $ 7.05 $ \\
\hline
\multirow{3}{*}{\rotatebox[origin=c]{90}{\textbf{Ours}}} &
Ours$^{rand}$ (\small{2.8\%})   & $ 77.2 $ & $ 88.62 $ & $ 3.7 $ \\
& Ours$^{i-1}$ (\small{2.8\%})   & $ \blu{85.87} $ & $ \blu{92.29} $ & $ \blu{1.83} $ \\
& Ours (\small{2.8\%})  & $ \textbf{86.88} $ & $ \textbf{93} $ & $ \textbf{1.12} $ \\

\bottomrule
\end{tabular}
}
\end{subtable}
\hfill
\begin{subtable}[h]{0.45\textwidth}
\centering
\caption{4 tasks VAW}
\label{tab:4task-vaw}
\resizebox{0.9\textwidth}{!}{
\begin{tabular}{c|l|c|c|c} %
\toprule
& \rule{0pt}{10pt} Method ($N_{prm}$) & Final Acc. ($\uparrow$) & $A_N$ ($\uparrow$) & $F_N$ ($\downarrow$) \\ %
\midrule
\multirow{1}{*}{\rotatebox[origin=c]{90}{VLPs}} & 
\multirow{2}{*}{BLIP \cite{blip}} & \multirow{2}{*}{$ 74.73  $} & \multirow{2}{*}{$ 78.3 $} & \multirow{2}{*}{$ - $} \\
& & &  \\
\hline
\multirow{8}{*}{\rotatebox[origin=c]{90}{Baselines}} &
CFT (\small{100\%}) & $ 80.15 $ & $ 83.88 $ & $ 7.41 $ \\
& CFT-FH (\small{99.7\%}) & $ 80.87 $ & $ 84.08 $ & $ 6.60 $ \\
& CFT-LP (\small{0.3\%}) & $ 65.46 $ & $ 74.98 $ & $ 15.28 $ \\
& L2 (\small{100\%})  & $ 83.72 $ & $ 86.01 $ & $ 2.95 $ \\
& LwF (\small{100\%})  & $ 81.20 $ & $ 85.38 $ & $ 6.26 $ \\
& EWC (\small{100\%})  & $ \blu{86.00} $ & $ 87.26 $ & $ \textbf{1.05} $ \\
& L2P (\small{$\le5\%$})  & $ 62.27 $ & $ 68.68 $ & $ 16.02 $ \\
& DP ((\small{$\le5\%$})  & $ 60.07 $ & $ 71.98 $ & $ 8.07 $  \\
\hline
\multirow{3}{*}{\rotatebox[origin=c]{90}{\textbf{Ours}}} &
Ours$^{rand}$ (\small{2.8\%})  & $ 85.43 $ & $ 88.78 $ & $ 1.4 $ \\
& Ours$^{i-1}$ (\small{2.8\%})  & $ \blu{86.00} $ & $ \blu{89.43} $ & $ 1.9 $ \\
& Ours (\small{2.8\%}) & $ \textbf{87.50} $ & $ \textbf{89.77} $ & $ \blu{1.19} $ \\

\bottomrule
\end{tabular}}

\end{subtable}
\tabvspace
\end{table*}

\secvspace
\noindent\textbf{Implementation Details.} Our code is built on top the official BLIP \cite{blip} repository. We start all compared methods (including ours) from BLIP pre-trained weights. Our $\mathcal{E}_I$ image encoder is ViTB/16 and our text encoder $\mathcal{E}_T$ is BERT with 12-layer encoder and 768 hidden size. As explained in \cref{sec:arch}, our decoder $\mathcal{D}$ is an extension of the text encoder $\mathcal{E}_T$, with cross-attention layers added between every two self-attention layers, each receiving $\mathcal{E}_I$-encoded image tokens as additional input. Our binary classifier $\mathcal{C}$ is a $2$-layer MLP with 768 hidden size. The entire model is comprised of 224M trainable parameters. All baselines including our method were tuned once on the VG \textit{color} $\rightarrow$ \textit{material} $\rightarrow$ \textit{size} attributes task sequence (i.e., they were tuned on a \emph{single} sequence in this particular order) using validation data, and afterward the same parameters were used throughout. We used WD 0.05 and an initial LR of 1e-5 and cosine scheduler in case of full model FT, and initial LR 1.25e-3 in all experiments involving the low-rank adapters (so in all instances of our proposed method). For our method, we used low-rank adapters rank $r=16$, $n_{adv}=10$, $\lambda_{adv}=0.01$, and $\rho=0.2$. Max of 12 epochs were used to train each task, in practice all compared methods converged earlier. For all the compared methods, for each task in the sequence, the val set of the current training task was used to automatically choose the best epoch.

\noindent\textbf{Metrics.} We employ 4 metrics to evaluate all models in all the experiments. The Final Accuracy is the accuracy of the final model (after training on all tasks in the sequence) evaluated on the test data of all the tasks without knowledge of task id. It is arguably the most interesting metric for practical applications. The $A_N$ averages the accuracy also over models corresponding to intermediate task sequences arising in the middle of the full sequence (it is commonly higher than final accuracy as it also averages over shorter and easier task sequences), $F_N$ provides the average forgetting~\cite{chaudhry2018efficient,Lopez-Paz:2017,hsu2018re} (average over performance drops of consecutive task models in the sequence), and finally $N_{prm}$ gives the \% of trainable parameters for each model.

\secvspace
\subsection{Comparison to Baselines}
\secvspace
For our main evaluations, we use the task sequences formed from the \svlc{} subsets defined for the VG, VAW, and VG+VAW datasets as part of our \oursbench{} benchmark as explained in \cref{sec:bench}. Specifically, we use the \textit{rel. spatial} $\rightarrow$ \textit{attr. size} $\rightarrow$ \textit{attr. material} $\rightarrow$ \textit{rel. action} $\rightarrow$ \textit{attr. color} $\rightarrow$ \textit{object state} $\rightarrow$ \textit{attr. action} $7$-task sequence on VG and VG+VAW, \textit{attr. action} $\rightarrow$ \textit{object state} $\rightarrow$ \textit{attr. color} $\rightarrow$ \textit{attr. size} 4-task sequence on VAW, and use \textit{attr. color} $\rightarrow$ \textit{attr. material} $\rightarrow$ \textit{attr. size} $\rightarrow$ \textit{rel. action} $\rightarrow$ \textit{object state} $\rightarrow$ \textit{attr. action} for our ablation study. The task order was set arbitrarily, and we evaluate the effect of tasks order in \cref{sec:ablation}. As VAW is a `clean' dataset, being comprised of pairs of explicit object and attribute annotations (as opposed to VG where annotations were collected in natural language form as unconstrained sentences), we make our VAW task sequence harder by excluding the \textit{attr. material} task, which has the highest performance in the original pre-trained BLIP that all compared methods use, as the starting point. The results of our method and comparisons to baselines for all the tested datasets and task sequences are provided in \cref{tab:7task} (VG+VAW), \cref{tab:7task-vg} (VG) and \cref{tab:4task-vaw} (VAW).

\begin{figure*}[h]
\figvspace
    \centering
    \begin{subfigure}{0.27\textwidth}
        \centering
        \includegraphics[width = \textwidth]{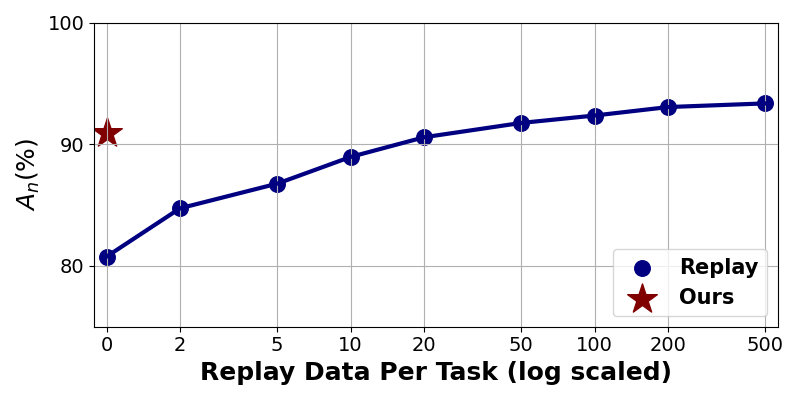}
        \caption{7 Task VG+Vaw}
    \end{subfigure}
    \begin{subfigure}{0.27\textwidth}
        \centering
        \includegraphics[width = \textwidth]{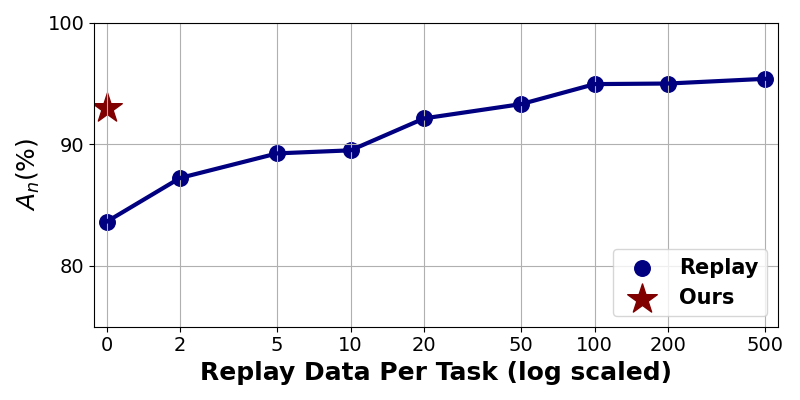}
        \caption{7 Task VG}
    \end{subfigure}
    \begin{subfigure}{0.27\textwidth}
        \centering
        \includegraphics[width = \textwidth]{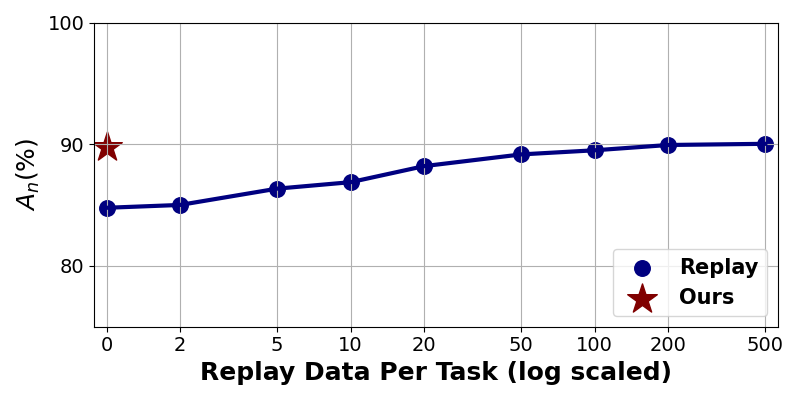}
        \caption{4 Task VAW}
    \end{subfigure}
    \vspace{-0.25cm}
    \caption{\textbf{Analysis of our approach versus replay of stored training data} with $A_N$ on the y-axis and data stored per task on the x-axis.}
    \label{fig-app:omega}
    \vspace{-5mm}
\end{figure*}
\noindent\textbf{\vl{} Pre-trained (VLP) models.} We evaluate multiple openly available VLPs, including the very popular CLIP \cite{clip} and BLIP \cite{blip} on our longest and largest VG+VAW task sequence in \cref{tab:7task}. We use their original implementations and pre-trained weights. All of these models are pre-trained on with millions of images, and yet, as noticed in \cite{winoground} and \cite{vlc}, without additional training are brittle with respect to understanding \svlc{}s which are the focus of our \oursbench{} benchmark. From \cref{tab:7task} we clearly see that BLIP has the highest out-of-the-box performance on our tasks among all the VLPs (and in general, being a more recent released model), hence we choose it as a starting point for all our training CL baselines as well as for our method. For the same reason, we use only BLIP to represent VLPs in smaller VG only and VAW only evaluations. Interestingly, 
naively fine-tuning BLIP in a continual setting (CFT baseline in \cref{tab:7task}) reduces its final accuracy below the original non-fine-tuned BLIP by 3.88\%. Also interestingly, our proposed approach improves the final accuracy of the CFT and the original BLIP by 12.28\% and by 8.4\% respectively. Similar strong gains in the final accuracy are observed for our method w.r.t. original BLIP and CFT in VAW and VG experiments in \cref{tab:4task-vaw} and in \cref{tab:7task-vg} respectively.

\noindent\textbf{Our model variants.} In all of \cref{tab:7task}, \cref{tab:7task-vg}, and \cref{tab:4task-vaw} we compare to 3 variants of our method. \textit{Ours} refers to our full approach detailed in \cref{sec:method}. \textit{Ours}$^{i-1}$ refers to a simplified version of our approach that performs the \oursloss{} w.r.t. only the previous task model, which in turn incorporates the knowledge of all previous tasks in a good way thanks to our proposed technique of reducing forgetting (apparent from the comparable performance of this variant compared to full approach). \textit{Ours}$^{rand}$ refers to a variant of our approach that employs the same \oursarch{} architecture and adversarial loss $\mathcal{L}_{adv}^j$ (defined in \cref{eq:adv_loss}), but generates the adversarial texts by random sampling of other texts in the batch. Although this approach obtains lower gains than our full method or \textit{Ours}$^{i-1}$, it still competes strongly with most of the baselines, highlighting the benefits of our \oursarch{} architecture and the proposed adversarial loss $\mathcal{L}_{adv}^j$.

\noindent\textbf{CL baselines comparison.} We compare to a wide variety of baselines (detailed in \cref{sec:baselines}), also including the very recent data-free visual CL SOTA prompting-based methods L2P \cite{l2p} and DP \cite{dualprompt} (using their respective methods PyTorch adaptation to \vl{}, which, as detailed in our \SupB{}, was verified to reproduce their vision-only results on their respective original papers benchmarks and was thoroughly tuned on the $3$ tasks same as all the compared methods).
\begin{table}[t]
\caption{\textbf{Ablations - 6 tasks VG+VAW}. Fin. Acc. is the performance of the final model on all tasks without task-id knowledge; $A_N$ is the accuracy averaged over tasks; $F_N$ is the average forgetting; all methods have $N_{param}=2.8\%$ trainable parameters.}
\label{tab:ablations}
\centering
\vspace{-2mm}
\resizebox{.42\textwidth}{!}{
\begin{tabular}{c|l|c|c|c} %
\toprule
& \rule{0pt}{10pt} Method & Fin. Acc. ($\uparrow$) & $A_N$ ($\uparrow$) & $F_N$ ($\downarrow$) \\
\midrule
\multirow{5}{*}{\rotatebox[origin=c]{90}{Ablations}} &
\oursarch & $ 73.68 $ & $ 78.7 $ & $ 6.21 $ \\
& LWF \oursarch & $ 70.83 $ & $ 74.86 $ & $ 5.72 $ \\
& Pos. Adv. & $ 76.3 $ & $ 82.15 $ & $ 5.8 $ \\
& Pos. Adv.$^{i-1}$  & $ 80.98 $ & $ 84.84 $ & $ 5.19 $ \\
& Adv. Image  & $ 81.2 $ & $ 86.12 $ & $ 4.22 $ \\
\hline
\multirow{3}{*}{\rotatebox[origin=c]{90}{\textbf{Ours}}} &
Ours$^{rand}$  & $ 82.71 $ & $ 86.97 $ & $ \blu{2.25} $ \\
& Ours$^{i-1}$  & $ \blu{84.69} $ & $ \blu{87.73} $ & $ 2.29 $ \\
& Ours  & $ \textbf{87.11} $ & $ \textbf{89.69} $ & $ \textbf{1.55} $ \\
\bottomrule
\end{tabular}
}
\vspace{-0.2cm}
\end{table}
As can be seen from \cref{tab:7task}, \cref{tab:7task-vg}, and \cref{tab:4task-vaw}, our method has significant advantages over all of the baselines, especially apparent in $\times5$ reduction in the forgetting metric $F_N$ and $6.8\%$ to $5.5\%$ gains in final accuracy (VG+VAW and VG) with respect to the strongest baseline. Smaller $1.5\%$ and $2.5\%$ gains in final accuracy and $A_N$ (w.r.t. strongest baseline) on VAW can likely be attributed to VAW being very clean data constructed from combining explicit object + attribute annotations, as opposed to VG whose annotations were collected as free-form natural language texts. Another interesting aspect is that our model uses significantly less parameters for adapting the base model, making it parameter efficient, and yet, as we show in our ablations \cref{sec:ablation}, the reduced number of trained parameters is clearly not the source of our methods greatly reduced forgetting. As revealed there, the source of reduced forgetting is certainly the proposed \oursloss{} technique building on our efficient \oursarch{} architecture.

\begin{table}[t]
\caption{\textbf{Task order ablation - mean over all 3-tasks orders with confidence intervals}. Final Accuracy measures the performance of the final model on all tasks without task-id knowledge; $A_N$ is the accuracy averaged over tasks; $F_N$ is the average forgetting.}
\vspace{-2mm}
\label{tab:orders}
\centering
\resizebox{.42\textwidth}{!}{
\begin{tabular}{c|c|c|c}%
\toprule
\rule{0pt}{10pt} Method & Fin. Acc. ($\uparrow$) & $A_N$ ($\uparrow$) & $F_N$ ($\downarrow$) \\
\midrule
CFT & $ 84.64 \pm 1.3 $ & $ 88.09 \pm 1.75 $ & $ 6.32 \pm 1.16 $ \\
\hline
Ours  & $ 89.6 \pm 0.29 $ & $ 90.56 \pm 1.32 $ & $ 1.23 \pm 0.27 $ \\
\bottomrule
\end{tabular}
}
\tabvspace
\end{table}

\secvspace
\subsection{Ablations}\label{sec:ablation}
\secvspace
To explore the different aspects of what makes the proposed approach successful, we use a reduced size VG+VAW sequence of $6$ tasks, removing the large \textit{rel. spatial} task for quicker experiments. We present the results of the ablation study in \cref{tab:ablations}, exploring the following variations of our method:
(i) \textbf{\oursarch{}} - evaluates our proposed \oursarchfull{} architecture on \oursbench{} without employing the proposed \oursloss{}, as we can see the reduced number of trained parameters in \oursarch{} alone is not sufficient to alleviate forgetting in \oursbench{}; 
(ii) \textbf{LWF \oursarch{}} - explores if the proposed \oursloss{} technique could be replaced by simple distillation from past models by applying LWF, as can be seen distillation does not help and even reduces the result due to reduced model plasticity; 
(iii) \textbf{Pos. Adv.} - simulates \textit{positive} pseudo-replay by reversing the adversarial gradient computed for \oursloss{} in our method and thus trying to generate samples more positive for past tasks, as we can see this significantly under-performs our proposed \oursloss{}, supporting the intuition that generating negatives for past tasks is much easier;
(iv) \textbf{Pos. Adv.$^{i-1}$} - \textit{positive} pseudo-replay could be significantly improved if only using single past task model, still strongly under-performing our \oursloss{}; 
Finally, (v) \textbf{Adv. Image} - verifies that generating adversarial texts in \oursloss{} is more beneficial then generating adversarial images, this is intuitive as an image is a more complex and less structured signal and hence harder to effectively manipulate.

We examine the effects of tasks order by evaluating our approach on \textit{all} possible task orders of \textit{color}, \textit{material}, and \textit{size} attributes task of VG+VAW, the results are presented in \cref{tab:orders}. Clearly, our method is very robust to task order with only minimal changes in the performance metrics.

Finally, although our proposed method and the \oursbench{} setting are data-free, it is still interesting to compare our results to increasing amounts of experience replay (prohibitive when data privacy and/or copyright need to be preserved) applied to the same sequences of tasks. We present this comparison in \cref{fig-app:omega}. We observe that our method is comparable to the highest amounts of experience replay, also improving over some replay amounts.

\secvspace
\section{Conclusions}
\label{conclusion}
\secvspace
We have presented a benchmark and a method for exploring a very practical and important task of continual improvement of \vl{} models in understanding \svlcfull{}. Our proposed approach builds upon the proposed efficient \oursarch{} architecture combined with the \oursloss{} technique to significantly reduce forgetting especially designed for \oursarch{} and the challenging multi-modal continual \vl{} setting. We have shown the advantages of the proposed approach w.r.t. strong CL baselines applied to the same setting demonstrating its effectiveness, as well as exploring different aspects of its design in our ablation studies and other experiments. Potential future work includes exploring applications of the proposed \oursarch{} and \oursloss{} techniques to uni-modal settings, as well as exploring potential additional improvements discussed in the text.

\section*{Acknowledgements}

This material is based upon work supported by the Defense Advanced Research Projects Agency (DARPA) under Contract No. FA8750-19-C-1001. Any opinions, findings and conclusions or recommendations expressed in this material are those of the author(s) and do not necessarily reflect the views of the Defense Advanced Research Projects Agency (DARPA).

{\small
\bibliographystyle{ieee_fullname}
\bibliography{references}
}

\appendix
\section*{Appendix}
\setcounter{figure}{0}
\setcounter{table}{0}
\renewcommand{\thetable}{\Alph{table}}
\renewcommand{\thefigure}{\Alph{figure}}
\renewcommand\thesection{\Alph{section}}

\section{Discussion on Choice of BLIP}
\label{blip-others}

We chose to base our method on BLIP as it had highest out-of-the-box performance (as a pre-trained model) on the ConStruct-VL tasks (Table 1) compared to numerous VL models including the very recent CyCLIP~\cite{cyclip}, thus making it a good representative source model for CL on ConStruct-VL. We also evaluated the out-of-the-box performance on ConStruct-VL tasks using METER~\cite{meter}, X-VLM~\cite{xvlm}, VLMO~\cite{vlmo}, and FIBER~\cite{fiber}, and observed an average performance of 56.8\%, 58.9\%, 54.6\%, and 73.9\% respectively (21.0\%, 18.9\%, 23.2, and 3.9\% below out-of-the-box BLIP), which further demonstrates the difficulty of VL models to understand VL concepts. As our approach is orthogonal to continued improvements in VL, we note that a great future direction is to explore future improved VL models with our approach on ConStruct-VL.

\section{Details on Prompting Baselines}
\label{appendix-prompt}

In \AppRefA{}, we discuss our PyTorch implementations of the very recent and influential L2P \cite{l2p} and DualPrompt \cite{dualprompt} works which are state-of-the-art (SOTA) data-free visual continual learning (CL) prompting-based methods. 
In our PyTorch implementation, we rigorously followed the description and the JAX code of L2P and DualPrompt. Furthermore, we tuned their hyperparameters for \oursbench{} by maximizing their performance on the same $3$ task sequence of \oursbench{} as for all of the compared methods, including all the baselines and our own approach (\AppRefB{}). 
In this section, we provide additional details on these L2P and DualPrompt baselines.

L2P and DualPrompt work by learning a key-value paired prompt pool based on an instance-wise query mechanism. For L2P, we use a prompt size of 4, prompt pool size of 50, and choose the 5 closest prompts from the pool at a time. For DualPrompt, we use a prompt length of 20 for the `expert' prompts, and a prompt length of 6 for the `general' prompts. Importantly, these hyperparameters were tuned in the same manner as for all other compared methods in our paper by maximizing performance on the same $3$ tasks sequence of \oursbench{} (starting from the hyperparameters recommended in the original papers \cite{l2p,dualprompt}). We also searched \emph{where} to insert prompts. Whereas originally L2P has prompting in layer 1 only, and DualPrompt has `general' prompts in layers 1,2 and `expert prompts' in layers 3,4,5; through tuning L2P and DualPrompt on \oursbench{}, we found that adding prompts in every layer of the model for both methods (i.e., layers 1-12 for L2P and layers 3-12 for DualPrompt `expert' prompts) maximizes their \oursbench{} performance.

We note that the under-performance of these methods on the proposed \oursbench{} benchmark \AppRefC{}
is likely an indication that the proposed problem of multi-modal continual learning of \svlc{}s in \oursbench{} is challenging to the vision-only CL SOTA and is thus an exciting new CL goal, which we just started to explore in the current work.

\end{document}